# National Institute on Aging PREPARE Challenge: Early Detection of Cognitive Impairment Using Speech — The SpeechCARE Solution

Maryam Zolnoori, PhD[1,2]; Hossein Azadmaleki, MS[1]; Yasaman Haghbin, MS[1]; Ali Zolnour, MS[1]; Mohammad Javad Momeni Nezhad, MS[1]; Sina Rashidi, MS[1]; Mehdi Naserian, MS[1]; Elyas Esmaeili, MS[1]; Sepehr Karimi Arpanahi, MS[1]

[1]Columbia University Irving Medical Center, New York, NY 10032, United States

[2]School of Nursing, Columbia University, New York, NY 10032, United States

**Corresponding Author**
Maryam Zolnoori, PhD
Columbia University Medical Center,
School of Nursing Columbia University
Phone: 317-515-1950
Address: 560 W 168th St, New York, NY 10032
E-mail: mz2825@cumc.columbia.edu; m.zolnoori@gmail.com

**Number of Tables:** 2

**Number of Figures:** 6

# A. ABSTRACT

Alzheimer's disease and related dementias (ADRD) affect one in five adults over 60, presenting a significant public health concern.[1–3] Despite nationwide efforts, over half of individuals with cognitive decline, including mild cognitive impairment (MCI) and ADRD, remain undiagnosed.[4–6] Recent studies showed that speech-based assessments hold promise for early detection: phonetic motor planning deficits impair vocal tract control (e.g., vocal folds), altering acoustic features like pitch and tone.[7] Memory and language impairments disrupt language organization, leading to syntactic and semantic errors and reduced fluency.[7,8,9] The studies mostly focused on conventional speech-processing pipelines with hand-crafted features (e.g., eGeMAPS[10]) or general-purpose audio classification models (e.g., YAMNet[11], VGGish[12]). However, these approaches often exhibit suboptimal performance and limited generalizability across diverse language settings. To address these gaps, we introduce SpeechCARE, a multimodal speech processing pipeline leveraging underutilized pretrained, multilingual acoustic and linguistic transformer[13] models to capture nuanced acoustic and linguistic cues associated with cognitive impairment. Inspired by Mixture of Experts (MoE) paradigm,[14] its core architecture is a novel multimodal fusion architecture that dynamically weights transformer-driven acoustic and linguistic features for effective integration, enhancing performance and generalizability across different speech production tasks (e.g. story recall, sentence reading). This fusion mechanism has the capability of seamless integration of additional data (e.g., social determinants,[15] MRI[16]), boosting its sensitivity across the cognitive impairment spectrum. By leveraging pretrained transformer models, SpeechCARE can tackle challenges posed by small sample sizes, thereby enabling the inclusion of often-overlooked linguistic diversity. SpeechCARE's robust preprocessing pipeline includes automatic transcription, LLM (Large Language Model[17])-assisted data anomaly detection, and LLM-assisted speech task identification. SpeechCARE's explainability framework visualizes each modality's contribution to decision-making, highlighting linguistic and acoustic cues linked to cognitive impairment through a novel SHAP-based approach and LLM-based reasoning. SpeechCARE achieved AUC = 0.88 and F1 = 0.72 in distinguishing cognitively healthy, MCI, and AD. Specifically, for MCI detection (versus control), AUC = 0.90 and F1 = 0.62. Bias analyses using equal opportunity and average odds showed no significant differences across demographics except for age over 80. Various techniques (e.g., oversampling, weighted loss) were used for bias mitigation. Future Directions. SpeechCARE is accessible, non-invasive, and cost-effective for real-world care settings. Our immediate goal is to validate its performance on patient-clinician communications from VNS Health and Columbia University's Alzheimer's Disease Research Center, focusing on underrepresented groups (Hispanic, Black) in New York City (NYC). With support from Columbia's visualization center, we aim to evaluate its explainability in clinician-centered design for Electronic Health Record (EHR) integration, enabling timely diagnosis of cognitive impairment and early intervention for diverse populations[18].

# B. METHODOLOGY

## B.1. Dataset characteristics

The PREPARE Challenge comprises 2,058 participants (1,646 in the training set and 412 in the test set), including 1,140 healthy controls, 268 individuals with mild cognitive impairment (MCI), and 650 individuals with Alzheimer's disease (AD) (Figure 1). Participants spoke three languages: English (1,655), Spanish (360), and Mandarin (43). Gender was reported as female or male, with females comprising 1,219. Ages ranged from 46 to 99 years (avg: 75.13 ± 8.65). Education was reported both numerically and categorically with 35.23% missing data. Race was reported for less than 8% of the participants (92.61% missing).

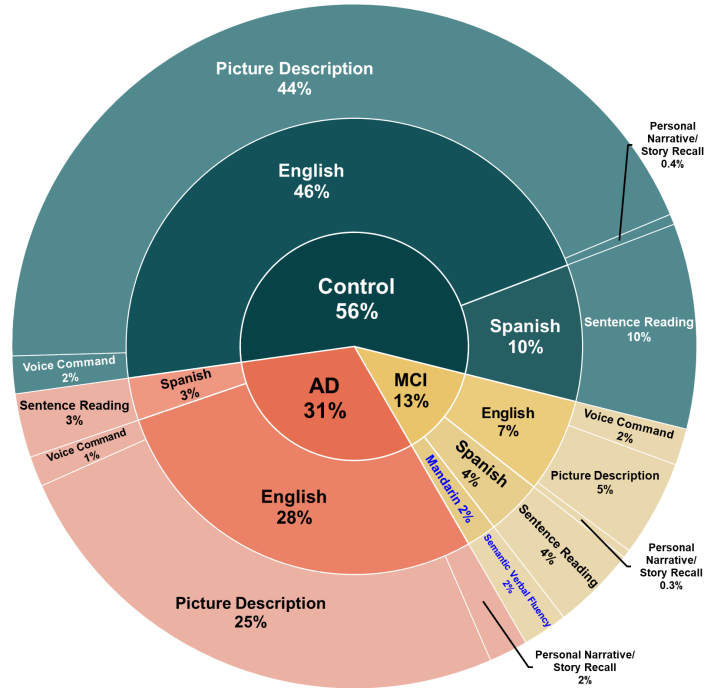

**Figure 1**. Statistics on diagnosis classes, language, and speech tasks.

### B.2. Preprocessing components

**(a) Demographic information preparation:** To facilitate model bias analysis and effectively incorporate demographic data, we grouped age into three categories following common study conventions: mid-life adults (46–65, 12%), older adults (66–80, 52.7%), and elderly (80+, 35.3%). To normalize education without missing values, we categorized it into four levels: no/elementary (11.61%), high school (24.98%), technical/undergraduate (16.72%), and advanced/graduate (11.47%). We addressed missing values (35.23%) using Iterative Imputer,[19] a multivariate regression method modeling them from age, gender, and language as input features.

**(b) Data quality evaluation and improvement:** High-quality audio data is essential for encoding subtle acoustic cues particularly for MCI detection.[20] However, the recordings included different types of stationary (e.g., recording device's hum) and dynamic noises (e.g., nearby conversations). To measure noisiness, we used different techniques including Signal-to-Noise Ratio and Spectral Flatness Measure, but due to inconsistent preprocessing by corpora developers, the outcomes were not strongly correlated with human perception. Therefore, we used a qualitative method, Mean Opinion Score,[21] where two experts (HA, MZ) listened to 5% of the data (stratified by demographic and diagnostic classes) and rated quality on a 1–5 Likert scale.[22] Findings showed 62.4% severely noisy (scores 1–2), 14.7% moderately noisy (score 3), and 22.9% slightly noisy (scores 4–5). To safely mitigate the noise, we then applied a low-pass filter[23] at 8,000 Hz to remove high-frequency noise while preserving human speech frequency components.

**(c) Automatic transcription:** To incorporate linguistic cues into the SpeechCARE pipeline, we utilized the latest version of Whisper-Large,[24] an open-source, multilingual automatic speech recognition system that achieved state-of-the-art (SOTA) performance on standard benchmarks. Whisper is able to detect the language of input speech. To evaluate Whisper's transcription accuracy, we used a stratified random sample (by demographic and diagnostic classes) comprising 5% of English, 10% of Spanish, and 15% of Mandarin audio data, ensuring sufficient sample sizes for each language. Transcriptions were compared to human-generated transcripts to calculate the word error rate[25] (WER): English = 0.12, Spanish = 0.27, and Mandarin = 0.32.

**(d) Speaker diarization:** SpeechCARE's diarization module builds on our previous study, which used a machine-learning algorithm to automatically separate the patient and clinician in recorded conversations. Based on our prior studies and initial findings of this study, incorporating available clinician data from certain corpora improved overall performance. Therefore, clinician speech was included in model training whenever available in the audio files of this dataset.

**(e) LLM-Assisted data anomaly detection:** During the evaluation of the Whisper transcriptions, we discovered that some audio files contained only the clinicians' speech, likely due to errors in speaker diarization performed by the corpus developers. To systematically identify these cases, we employed LLaMA 405B[26]—an open-source LLM comparable to GPT-4 performance—with prompt engineering[27] and few-shot learning[28], guided by iterative analysis of the model's output. This approach revealed 22 audio files consisting solely of clinicians' speech. We removed these files from the training process to prevent potential confusion in the model.

**(f) LLM-assisted speech task identification:** Speech production task type was not available in the dataset, yet identifying it is important for model development and bias/error analysis. Each task requires distinct cognitive effort and yields different acoustic and linguistic cues for cognitive impairment detection. Reflecting broader trends in clinical NLP, LLMs are increasingly used in healthcare for data quality, task automation, and clinical text analysis[29]. We used LLaMA 405B with prompt engineering and few-shot learning, guided by iterative expert analysis (HA, MZ), to refine the LLM prompt for task identification. The same experts then verified 10% of the assigned tasks (stratified by demographics and diagnostic classes), confirming 98% accuracy. Errors arose mainly from truncated or insufficient transcripts. See Figure 1 for statistics on the identified tasks.

**(g) Task homogeneity and bias introduction:** According to these analyses, we observed that (1) all Spanish speakers performed only the sentence reading task, which may not provide sufficient temporal cues linked to cognitive impairment compared to picture description and story recall. This limitation introduces bias and may reduce pipeline performance for these participants; (2) all Mandarin speakers had MCI in both the training and test sets, inflating speech pipeline performance because the system simply learns to classify all Mandarin speakers as having MCI.

### B.3. Building speech processing models

Our team has extensive experience developing speech processing pipelines for cognitive impairment detection, highlighted in news media[30–33] and high-impact journals.[34–45] Overall, no single approach consistently yields optimal performance due to variations in speech corpus characteristics (e.g., speech task, language) and audio data quality. Consequently, we investigated various linguistic and acoustic processing methods for this dataset to build SpeechCARE pipeline on the top-performing and most generalizable models (See Table 1 for brief information and results).

**(a) Hand-crafted speech processing models:** We developed three pipelines based on hand-crafted acoustic and linguistic features (see more information and results in Table 1): (1) SpeechDETECT:[34] Drawn from a systematic review of studies focused on speech and language for MCI detection, we built SpeechDETECT, providing a comprehensive, explainable set of acoustic and temporal features of speech; (2) eGeMAPS-Based model: We built this model on the eGeMAPS V02[46] parameter set (provided by the challenge organizer), using advanced time-series analysis;[13] (3) LinguisticDetect:[42] Drawn from previous speech-related studies for MCI detection, we built this model by quantifying lexical richness, syntactic complexity, semantic coherence, and psychological cues linked to cognitive impairment using natural language processing methods.

**(b) Speech (acoustic) and linguistic pre-trained transformer models:** Pre-trained speech transformer models (e.g., HuBERT,[47] Wav2Vec 2.0[48]) are powerful in capturing long-term acoustic and temporal dependencies. However, they have been underutilized in cognitive impairment related studies due to the lack of effective fine-tuning procedure on small speech datasets. For this dataset, we evaluated (1) SOTA pre-trained speech-transformer models—Wav2Vec2 XLS-R,[49] mHuBERT;[50] (2) pre-trained linguistic-transformer models—mBERT[51] and XLM-RoBERTa[52] (commonly used models) as well as mGTE[53] and BGE[54] (SOTA models). We selected these models due to multilingual pre-training and high generalizability across diverse languages. Our initial analysis on this dataset showed that multilingual models (e.g., mHuBERT, mBERT) generally outperformed monolingual counterparts (HuBERT, BERT[55]), even for English-speaking participants; (3) Audio Spectrogram Transformer[56] (AST). SOTA Speech transformer models primarily were designed for automatic speech recognition. Therefore, we also tested the AST model, a common model for audio classification trained on YouTube audio. All these transformer models were fine-tuned using the procedure and architecture described in section B.5.

**(c) LLM-based models:** We investigated LLMs, which extend transformer architectures by greatly increasing model parameters and training data size. Prior work showed that LLMs excel at recognizing linguistic cues linked to cognitive impairment with minimal fine-tuning.[57–59] For local, HIPAA-compliant deployment, we fine-tuned: LLaMA 3[26] (versions 3.1 70B, 3.1 8B, 3.2 3B) and Ministral 8B[60] using QLoRA[61] (a parameter-efficient, low-rank adaptation method) combined with a linear classifier on the final-layer embeddings to generate probabilities.

### B.4. Evaluating speech processing models

We evaluate the performance of the processing pipelines by (a) creating a validation dataset by randomly selecting 20% of the training dataset, stratified using demographic and diagnostic classes—a standard practice in deep learning algorithms development; (b) standard metrics for multi-class evaluation: AUC-ROC (Area Under the Curve - Receiver Operating Characteristic, one-vs-rest approach for three classes), F1-score (harmonic mean of precision and recall) using the micro-average, which accounts for class imbalance, and multi-class log loss as instructed by challenge organizers. See Table 1 for performance of the speech processing models on the validation set.

**Table 1.** Summary of acoustic and linguistic models and their performance

| Models Characteristic | | | | Validation Result | | |
|---|---|---|---|---|---|---|
| **1.1. Acoustic Hand-Crafted Features** | | | | | | |
| *Models* | *Description* | | | *Loss* | *AUC* | *F1* |
| SpeechDetect | 6848 features encoding Frequency, Spectral, Voice Quality, Intensity, Signal Complexity, Rhythmic Structure, Fluency, and Speech Dynamics aspects of speech | | | 0.801 | 0.80 | 0.62 |
| eGeMAPS | 88 features encoding Frequency, Spectral, Voice Quality, and Intensity aspects of speech | | | 0.879 | 0.77 | 0.57 |
| **1.2. Linguistic Hand-Crafted Features** | | | | | | |
| *Model* | *Description* | | | *Loss* | *AUC* | *F1* |
| LinguisticDETECT | N features encoding lexical density, syntactic structure, semantic and psycholinguistic cues | | | 0.900 | 0.70 | 0.63 |
| **1.3. Speech Transformer Models** | | | | | | |
| *Transformer* | *Size* | *Training source* | *input* | *Loss* | *AUC* | *F1* |
| Wav2vec2 XLS-R | 300 M | 436K hours /128 languages | Waveform | 0.849 | 0.78 | 0.56 |
| mHuBERT | 96 M | 90K hours /147 languages | Waveform | 0.738 | 0.84 | 0.66 |
| AST | 87 M | 2M 10-second audio clips | Spectrogram | 0.803 | 0.80 | 0.60 |
| **1.4. Linguistic Transformer Models** | | | | | | |
| *Transformer* | *Size* | *Training source* | | *Loss* | *AUC* | *F1* |
| XLM-RoBERTa | 600 M | ~295 billion tokens across 100 languages | | 0.841 | 0.80 | 0.64 |
| BGE | 567 M | ~1.2 billion text pairs across 194 languages | | 0.806 | 0.81 | 0.66 |
| mGTE | 305 M | ~1028 billion tokens across 75 languages | | 0.789 | 0.83 | 0.67 |
| mBERT | 177 M | ~2.5 billion tokens across 104 languages | | 0.803 | 0.82 | 0.66 |
| **1.5. Large Language Models** | | | | | | |
| *LLM* | *Size* | *Availability* | *Fine-Tuning Method* | *Loss* | *AUC* | *F1* |
| LLaMA 3.1 | 70 B | Open weight | QLoRA | 0.787 | 0.82 | 0.68 |
| LLaMA 3.2 | 3 B | Open weight | QLoRA | 0.855 | 0.80 | 0.65 |
| LLaMA 3.1 | 8 B | Open weight | QLoRA | 0.811 | 0.81 | 0.66 |
| Ministral | 8 B | Open weight | QLoRA | 0.817 | 0.81 | 0.64 |

**B.5. Feature Network**

The feature network generates modality-specific representations from pre-trained linguistic, acoustic, and demographic inputs, which collectively form the basis of the multimodal fusion architecture (Figure 2).

Based on the comparative evaluation in Sections B.3 and B.4, we selected mGTE and mHuBERT as the core components of the SpeechCARE feature network.

**(a) Linguistic Transformer (mGTE).** Pre-trained linguistic transformers have been used in prior studies of cognitive impairment with varying results depending on data size, speech task, and language coverage. In SpeechCARE, the multilingual Generative Text Encoder (mGTE) (an encoder-only transformer comprising approximately 305 million parameters) was employed due to its large multilingual pretraining corpus (≈ 1,028 billion tokens across 75 languages) and extended context window (8,192 tokens), which far exceeds that of BERT (512 tokens). This extended context facilitates the modeling of syntactic errors, lexical disruptions, and semantic incoherence commonly observed in mild cognitive impairment and dementia. For each tokenized transcript sequence $T = \{t_1, ..., t_n\}$, mGTE produces contextual embeddings $h_i$. The embedding corresponding to the classification token, $h_{[CLS]}$, serves as the summary representation of the entire utterance. During fine-tuning, $h_{[CLS]}$ is projected through a fully connected layer with Tanh activation to yield a diagnostic probability vector across cognitive classes (Control, MCI, AD). Our empirical results showed that the linguistic-only model (mGTE) achieved a strong standalone performance with a test F1-score of 68.88%, confirming its capacity to extract rich lexical and syntactic features from transcripts.

**(b) Acoustic Transformer (mHuBERT).** Recent self-supervised acoustic encoders such as Wav2Vec 2.0 and HuBERT achieve state-of-the-art performance on various speech benchmarks but face challenges when applied to long clinical recordings. Their computational cost grows quadratically with sequence length, and they lack a global embedding summarizing temporal dependencies. To address these limitations, SpeechCARE adopts mHuBERT, a multilingual variant of HuBERT with 98 million parameters pre-trained on 90,000 hours of speech in 147 languages. Each 30-second waveform x is divided into 5-second segments with 25% overlap, producing up to seven overlapping windows. Every segment $x_i$ is encoded into 250 frame-level embeddings (≈ 25 ms windows). The concatenated segment embeddings form a unified sequence of up to 1,750 vectors representing the full audio. A learnable [CLS] embedding is prepended to this sequence[62], which is subsequently processed by a Customized Self-Attention Encoder (CSE) composed of two stacked attention blocks with four heads each, followed by normalization, dropout, and residual connections. This lightweight encoder efficiently aggregates temporal information across segments while preventing overfitting on a relatively small dataset. The resulting global acoustic representation $h_{A,[CLS]}$ encapsulates spectral, prosodic, and temporal attributes associated with cognitive decline. A feed-forward classification layer then maps this vector to diagnostic logits. Our ablation studies on acoustic modeling showed that stepwise architectural enhancements consistently improved performance: adding a learnable [CLS] embedding to the base mHuBERT model increased the F1-score from 66.80% to 67.77%, and further incorporating segmentation raised it to 68.23% (see Table 2).

To evaluate the influence of preprocessing, a variant termed SpeechCARE-AGF: Raw Audio was additionally trained using the same architecture without any noise reduction. An alternative configuration, SpeechCARE-AGF: CMGAN-Enhanced Audio, employed the Conformer-based Metric-GAN (CMGAN) model for neural denoising. As summarized in Table 2, both variants yielded lower F1-scores compared with the final configuration, indicating that unfiltered or excessively enhanced speech may distort cognitively relevant acoustic cues. Finally, a low-pass filter at 8 kHz was applied to the raw audio to suppress high-frequency noise while preserving the human voice frequency range. This preprocessing step, corresponding to the SpeechCARE-AGF: Low-Pass Filtered Audio variant in Table 2, produced the best results and was adopted for the final model configuration.

**(c) Demographic Representation.** Demographic attributes (age, gender, education) were encoded categorically via one-hot representation and projected through a dense layer to produce a compact latent vector $h_D$. Preliminary analyses indicated that categorical encoding of age yielded better discrimination than continuous values. The resulting vector was incorporated as a third modality within the fusion network. While demographics alone yielded lower performance (F1 = 55.70%), our evaluation showed that age contributed the most value among the demographic features and was therefore retained in the final fusion configuration.

**(d) Training and Hyperparameter Configuration.** Both mHuBERT and mGTE encoders were fine-tuned jointly within a unified multimodal architecture using separate learning rates ($1 \times 10^{-5}$ for mHuBERT and $1 \times 10^{-6}$ for mGTE) and a weight decay of $1 \times 10^{-3}$. The batch size was set to 4, and the CSE module included two attention blocks with four heads each. All fully connected layers comprised 128 neurons with Tanh activation and a dropout rate of 0.1. This configuration provided stable optimization while preventing overfitting on the limited training data available in the PREPARE dataset.

## B.6. Fusion Network

Multimodal integration is essential for capturing complementary cues across acoustic, linguistic, and demographic domains. Four fusion paradigms were evaluated (Intermediate Fusion, Scaled Late Fusion, Cross-Modal Attention, and Adaptive Gating Fusion (AGF)) to determine the most effective strategy (see Table 2). These fusion strategies were selected based on prior studies that compared early, intermediate, and late fusion methods in multimodal deep learning systems[63–65]. The Adaptive Gating Fusion model (Figure 2) was inspired by the Mixture-of-Experts (MoE) paradigm[14, 66] and designed to dynamically modulate the contribution of each modality. It achieved the highest validation and test performance and was therefore adopted as the final architecture.

**(a) Adaptive Gating Fusion (AGF) Architecture.** The Adaptive Gating Fusion model (Figure 2) was inspired by the Mixture-of-Experts (MoE) paradigm and designed to dynamically modulate the contribution of each modality. Given feature embeddings from the feature network—acoustic ($\boldsymbol{h_A}$), linguistic ($\boldsymbol{h_L}$), and demographic ($\boldsymbol{h_D}$)—AGF computes a weighted combination of modality-specific outputs through three sequential stages:

1. Modality encoding: Each $h_i \in \{h_A, h_L, h_D\}$ passes through a fully connected layer with Tanh activation, yielding hidden representations $z_i$.

2. Gating network: The concatenated vector $[z_A; z_L; z_D]$ is passed to a gating subnetwork consisting of a single fully connected layer followed by a Softmax function that produces attention weights $w_i$ for each modality ($\sum_i w_i = 1$).

3. Weighted aggregation: Modality-specific prediction scores $s_i$ are computed via separate linear classifiers and combined into the final output:

$$y = \sum_i w_i s_i$$

This procedure adaptively emphasizes modalities contributing the most informative cues for each input. Furthermore, this gating mechanism provides interpretable insights into modality reliance (see Table 2) and enhances robustness when a modality is degraded or missing.

**(b) Comparative Performance and Statistical Validation.** As shown in Table 2, the AGF configuration integrating acoustic (mHuBERT + CSE), linguistic (mGTE), and age as the demographic variable achieved the best overall results on the PREPARE test set (AUC = 86.83 ± 0.46 %, F1 = 72.11 ± 0.44 %). This performance exceeded that of all other fusion strategies— Intermediate Fusion (AUC = 86.28%, F1 = 70.10%), Scaled Late Fusion (AUC = 86.21%, F1 = 70.29%), and Cross-Modal Attention (AUC = 86.61%, F1 = 70.51%). In addition, when substituting other demographic features, performance decreased: using education or gender alone resulted in F1-scores of 69.20% and 69.95% respectively, while incorporating all demographics together yielded an F1 of 68.42%. To assess the significance of these differences, paired t-tests confirmed that improvements achieved by AGF over competing fusion approaches were statistically significant ($p < 0.01$) with large effect sizes (Cohen's $d > 1.5$). Furthermore, among the baseline fusion strategies, Cross-Modal Attention slightly outperformed Intermediate and Scaled Late Fusion, but AGF consistently delivered ~1.5–2.0% higher F1 and retained top AUC performance across configurations. Taken together, these findings establish AGF as the optimal framework for integrating multimodal representations in the SpeechCARE architecture.

**(c) Interpretation and Advantages.** The AGF framework offers several advantages:

1. Dynamic adaptation: weights vary across tasks and languages, enabling sensitivity to cognitive and structural demands.
2. Interpretability: learned weights indicate relative contributions of linguistic and acoustic cues.
3. Efficiency: the gating layer adds minimal computational overhead while maintaining competitive AUC–F1 trade-offs.
4. Robustness: adaptive weighting mitigates performance loss under noisy or missing modalities.

Overall, the combination of the Feature Network (mHuBERT + CSE and mGTE) and the AGF Fusion Network (Figure 2) yielded the top-performing configuration summarized in Table 2, demonstrating the effectiveness of transformer-based multimodal modeling for cognitive impairment detection.

**Table 2.** Comparative and ablation results for SpeechCARE model components

| Model | Validation | | Test | |
|---|---|---|---|---|
| | AUC | F1-Score | AUC | F1-Score |
| **Acoustic-Only Refinements** | | | | |
| mHuBERT (Base Model) | 84.03±1.09 | 66.78±2.27 | 84.07±0.60 | 66.80±1.25 |
| mHuBERT + CLS Embedding | 84.92±1.23 | 68.06±1.83 | 84.99±0.60 | 67.77±1.06 |
| mHuBERT + CLS Embedding + Segmentation | 84.55±1.14 | 67.60±1.56 | 84.85±0.70 | **68.23±1.11** |
| **Single-Modality Baselines and Modalities Integration** | | | | |
| All Demographics (Age, Gender, Education) | 72.78±0.79 | 55.82±0.80 | 72.31±0.71 | 55.70±0.43 |
| Voice (mHuBERT + CLS Embedding + Segmentation) | 84.55±1.14 | 67.60±1.56 | 84.85±0.70 | 68.23±1.11 |
| Transcription (mGTE) | 81.26±1.17 | 63.70±1.26 | 85.00±0.40 | 68.88±0.78 |
| Fusion-AGF: Voice + Transcription | 84.42±1.95 | 67.57±2.36 | 86.57±0.45 | 70.51±0.93 |
| Fusion-AGF: Voice + Transcription + All demographics | 83.49±0.96 | 66.14±1.74 | 85.49±0.69 | 68.42±0.78 |
| Fusion-AGF: Voice + Transcription + Education | 83.19±2.09 | 66.02±2.87 | 85.99±0.68 | 69.20±1.00 |
| Fusion-AGF: Voice + Transcription + Gender | 84.02±1.67 | 66.78±2.18 | 86.35±0.45 | 69.95±0.68 |
| Fusion-AGF: Voice + Transcription + Age | 84.97±1.57 | 68.12±2.69 | 86.83±0.46 | **72.11±0.44** |
| **Fusion Strategies** | | | | |
| Intermediate Fusion | 85.07±1.35 | 67.94±2.09 | 86.28±0.48 | 70.10±0.78 |
| Scaled Late Fusion | 85.19±1.38 | 68.62±2.77 | 86.21±0.57 | 70.29±1.13 |
| Cross-Modal Attention (+ Intermediate Fusion) | 85.49±1.50 | 68.58±2.05 | 86.61±0.56 | 70.51±0.71 |
| Adaptive Gating Fusion (AGF) | 84.97±1.57 | 68.12±2.69 | 86.83±0.46 | **72.11±0.44** |
| **Noise Reduction (Audio Preprocessing)** | | | | |
| SpeechCARE-AGF: Raw Audio | 84.25±1.33 | 85.76±0.80 | 85.76±0.80 | 69.15±1.01 |
| SpeechCARE-AGF: CMGAN-Enhanced Audio | 83.22±1.89 | 65.34±2.26 | 85.80±1.10 | 69.00±1.12 |
| SpeechCARE-AGF: Low-Pass Filtered Audio | 84.97±1.57 | 68.12±2.69 | 86.83±0.46 | **72.11±0.44** |

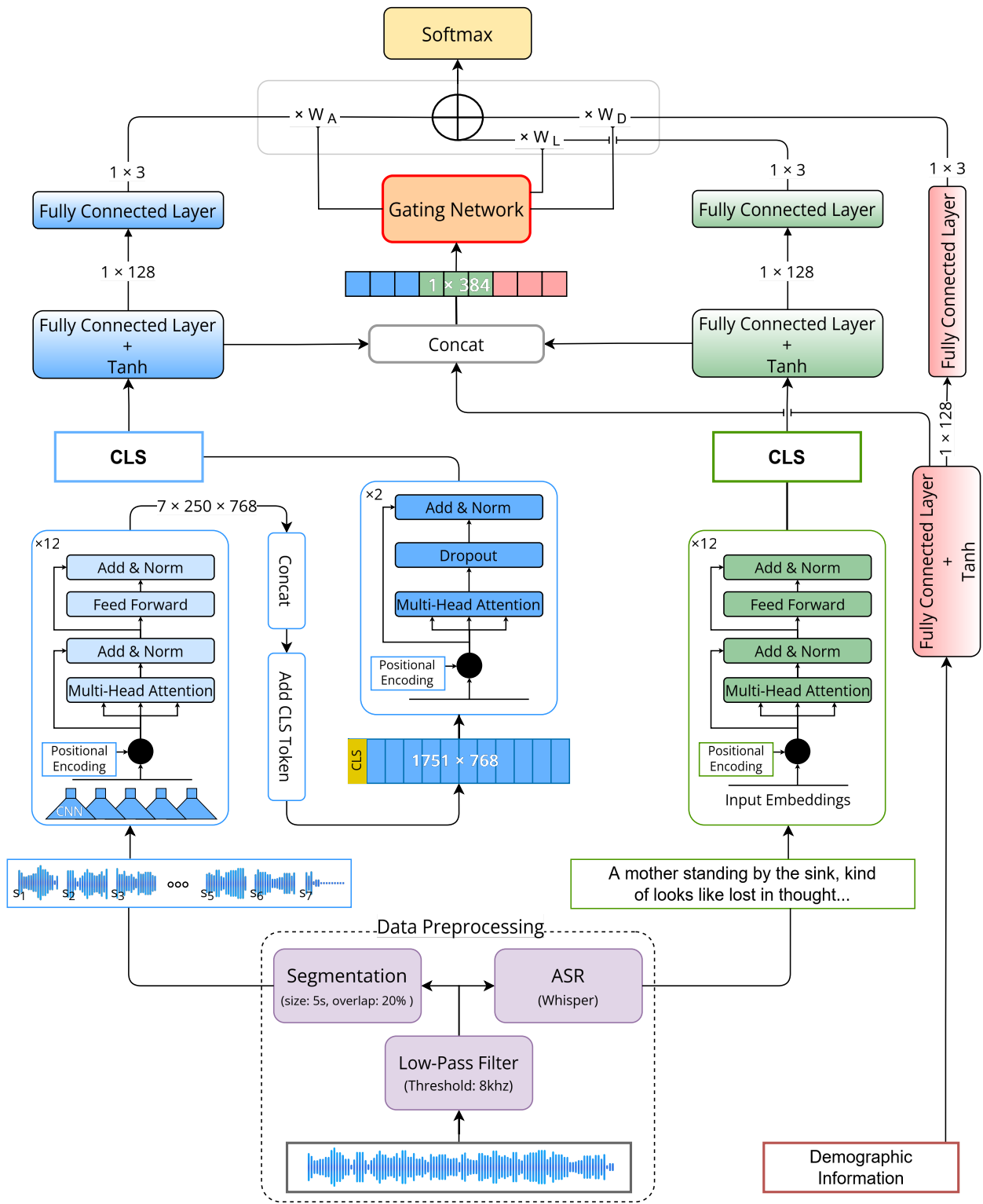

**Figure 2.** Architecture of the SpeechCARE Feature and Fusion Networks. The diagram illustrates how raw audio (up to 30 s) is processed via low-pass filtering (8 kHz cutoff) and segmentation (5 s chunks with 25 % overlap), followed by automatic speech recognition (ASR) to produce text. The acoustic pathway uses a multilingual HuBERT encoder plus a Customized Self-Attention Encoder (CSE) to extract a global [CLS] embedding, while the linguistic pathway processes transcripts with mGTE. Demographic data enter as a third modality. Each modality's embedding passes through its own dense layer, and the Adaptive Gating Network fuses these representations to generate final predictions.

## C. PERFORMANCE ANALYSIS

**(a) Performance metrics.** We used AUC-ROC (one-vs-rest approach) to evaluate how well the model distinguishes each class. We computed micro-average and weighted-average AUC: micro-average aggregates decisions across all classes equally, while weighted-average accounts for class imbalance. We also calculated Precision-Recall (PR) curves using the same approach. Additionally, we introduced Information Gain curves to measure how much uncertainty the model's predictions reduce compared to random guessing (Figures 3.A–3.C).

**(b) Performance of SpeechCARE's models** (Figures 3.A–3.C). **(a) SpeechCARE-AGF** with adaptive gating fusion (AGF) on mHuBERT, mGTE, and demographics achieved AUC=87.59 (Micro) and PR=77.17 (Micro). The cumulative gain curve showed that by targeting the top 40% of the population, ~80% of True diagnoses were captured; **(b) SpeechCARE-Whisper** replaces mHuBERT with Whisper-Medium (using encoder embeddings). Overall, Whisper dominated mGTE and demographics, causing the model to assign minimal/zero weight to them. This likely happened because Whisper was trained on a larger labeled dataset (680k h) compared to mHuBERT's smaller unlabeled set (90k h). Nonetheless, SpeechCARE-Whisper's performance did not significantly differ from SpeechCARE-AGF, indicating the robust AGF architecture for speech processing.

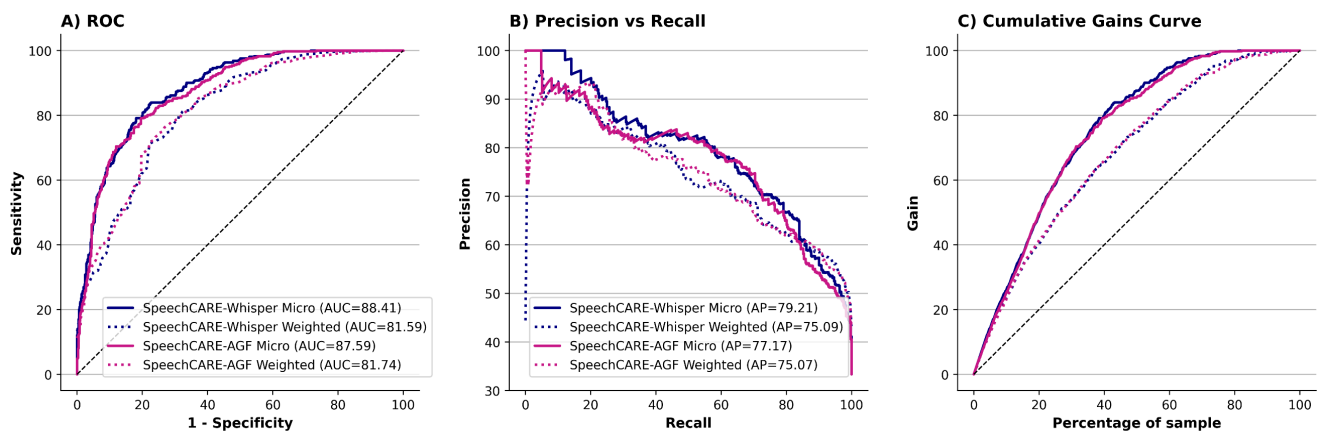

**Figure 3**. SpeechCARE-AGF & Whisper

**(c) Performance of SpeechCARE-AGF and SpeechCARE-Whisper for MCI detection (Figures 4.A–4.C).** Similarly to C.2, there was no significant difference between both models for MCI detection (MCI class vs. Control class). Overall, the result is promising with AUC-ROC = 90.50, RP = 68. **After threshold optimization,** we achieved F1= 0.62, recall=0.73, precision=0.54 for MCI, indicating a good performance of SpeechCARE for early detection.

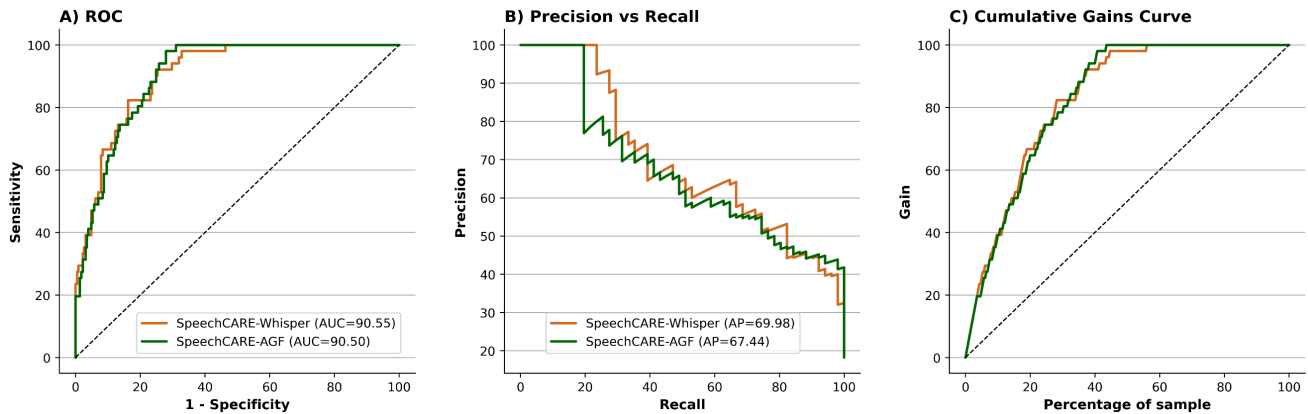

Figure 4. SpeechCARE-AGF & Whisper: MCI vs. Control

**(d) Error analysis (Figure 5).** We observed no false negatives (misclassified MCI/AD as controls) in Mandarin speakers because all Mandarin speakers in the dataset had MCI. Spanish speakers showed the highest false negatives, likely reflecting the limited ability of sentence-reading tasks to capture acoustic/linguistic cues (see B.2.(f)). English speakers primarily completed picture-description and narrative tasks. For these participants, we used a T-test (p=0.05) to compare acoustic (fundamental frequency, jitter, shimmer, intensity, MFCC) and linguistic (lexical richness, repetition, fillers, sentence structure) features between false negative class and control class (true negative). No significant differences were found in either feature set between these two groups. This may indicate that language-related brain regions may be less affected in the false-negative group. Incorporating additional modalities (e.g. medical history, MRI) may improve SpeechCARE sensitivity.

|  | Control | MCI | AD |
|---|---|---|---|
| **Control** | 190 | 30 | 9 |
| **MCI** | 10 | 41 | 0 |
| **AD** | 47 | 19 | 66 |

Figure 5. SpeechCARE-AGF Confusion Matrix. Note that the Thresholds were adjusted for three classes.

### D. BIAS ANALYSIS

**(a) Metrics.** We used Equality of Opportunity[67] (EOO) and Average Odds, two of the most common bias metrics, as widely adopted in recent evaluations of ASR fairness[68]. For demographic data (**Figure 6.A**), no significant bias was observed except for the age-over-80 group. For languages (**Figure 6.A**), bias stemmed from dataset constraints: (1) all Mandarin speakers had only MCI, inflating EOO; (2) all Spanish speakers only completed sentence-reading tasks, limiting critical speech cues for detection.

**(b) Methods.** We applied oversampling as a preprocessing method, incorporating Frequency Masking[69] and Voice Conversion[70] for speech augmentation, alongside in-processing methods such as Adversarial Debiasing[71] and reweighting[72]. For Spanish speakers, SpanBERTa (a Spanish transformer) was replaced with mGTE.

**(c) Results of mitigation.** Among oversampling techniques, frequency masking outperformed voice conversion, consistent with SpeechCura[73], which found both methods effective for improving cognitive impairment detection under limited data. Focal loss slightly reduced biases, while other in-processing methods showed no improvement. Replacing SpanBERTa with mGTE significantly mitigated bias. Overall, these methods improved EOO by 5.17% for the age-over-80 group and by 16.36% for Spanish speakers (**Figure 6.B**), demonstrating the effectiveness of our bias mitigation approach.

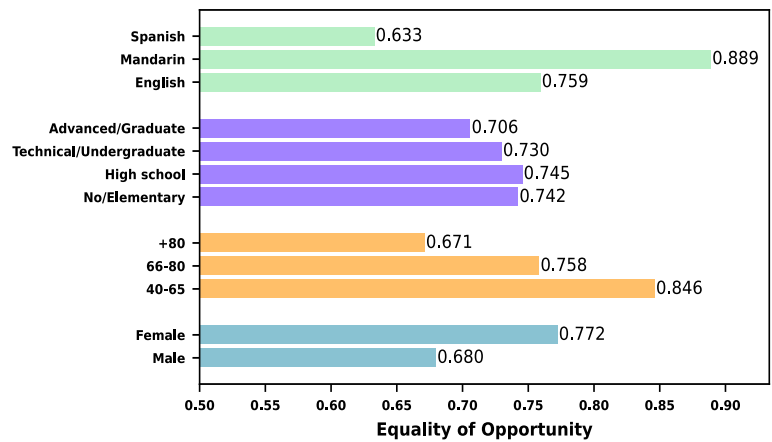

**Figure 6.A.** Equality of Opportunity before bias mitigation

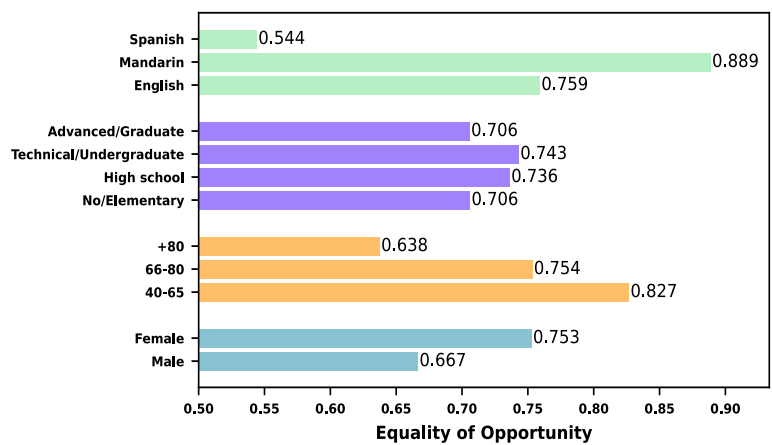

**Figure 6.B.** Equality of Opportunity after bias mitigation

**E. FUTURE DIRECTIONS**

**(a) Multimodal integration of speech, biomarkers, EHR data, social determinants from Columbia Alzheimer's Disease Research Center (ADRC).** Speech-only markers may underperform when language-related brain regions are less affected. We are collaborating with Columbia ADRC to collect speech data from racially diverse groups in NYC. We plan to (1) fine-tune SpeechCARE on the multilingual dialects, (2) apply SpeechCARE's adaptive gating fusion to measure speech's added value for early detection, (3) co-train speech embeddings with imaging biomarkers to uncover cross-modal links between speech features and brain changes.

**(b) Fine-tuning SpeechCARE on routine patient-clinician communication and enhancing explainability framework.** Speech production tasks overlook vital communication cues. We are collaborating with VNS Health (the largest home healthcare agency in the U.S.) to collect routine patient–clinician communications from a racially diverse population. We plan to (1) fine-tune SpeechCARE on the communications data and (2) refine SpeechCARE's Explainability Framework in a clinician-centered design, preparing it for efficacy trial and EHR integration.

**(c) Longitudinal monitoring of speech.** A single speech recording overlooks gradual cognitive decline. In partnership with Columbia Data Visualization Center, we are developing "SpeechCARE Lite," a mobile app for recording speech samples over time. We plan to (1) integrate time-series models (e.g., temporal transformers) for longitudinal analysis; (2) improve SpeechCARE Explainability to provide insights on cognitive changes over time.

**(d) Improving pre- and in-processing components of SpeechCARE.** Speaker diarization errors, transcription bias, and noise in audio data remain problematic. We plan to (1) use advanced deep learning methods for noise reduction, (2) fine-tune Whisper to reduce bias in racially diverse populations, (3) develop LLM-based speaker diarization, and (4) extend SpeechCARE's fusion model as larger datasets become available.

**Author Contributions:**

- Maryam Zolnoori: Speech processing Lab; Lead project

- Hossein Azadmaleki: Development of the SpeechCARE model; Project management

- Yasaman Haghbin: Fine-tuning Transformer models; Bias analysis; Error analysis

- Ali Zolnour: Performance analysis; Results visualization

- Mohamad Javad Momeni Nezhad: Fine-tuning Large language models

- Sina Rashidi: Using data augmentation techniques, Signal processing

- Mehdi Naserian: Building machine learning algorithms on handcrafted acoustic features

- Elyas Esmaeili: Building machine learning algorithms on handcrafted acoustic features

- Sepehr Karimi Arpanahi: Using LLM for task description

**Data Availability**

The dataset employed in this study was provided exclusively to our team as part of our participation in the 2024 NIA PREPARE Challenge. This dataset is not publicly available, and data sharing is restricted under the challenge terms. Researchers seeking access may contact the challenge organizers [here]. As the dataset has since been integrated into DementiaBank, access may alternatively be requested through the DementiaBank administrative team.

## F. REFERENCES

1. Busse, A., Hensel, A., Gühne, U., Angermeyer, M. C. & Riedel-Heller, S. G. Mild cognitive impairment: long-term course of four clinical subtypes. *Neurology* 67, 2176–2185 (2006).

2. Ganguli, M. *et al.* Prevalence of mild cognitive impairment by multiple classifications: The Monongahela-Youghiogheny Healthy Aging Team (MYHAT) project. *The American Journal of Geriatric Psychiatry* 18, 674–683 (2010).

3. Gagnon-Roy, M. *et al.* Preventing emergency department (ED) visits and hospitalisations of older adults with cognitive impairment compared with the general senior population: what do we know about avoidable incidents? Results from a scoping review. *BMJ Open* 8, (2018).

4. Boise, L., Neal, M. B. & Kaye, J. Dementia assessment in primary care: results from a study in three managed care systems. *J Gerontol A Biol Sci Med Sci* 59, M621–M626 (2004).

5. Tóth, L. *et al.* A speech recognition-based solution for the automatic detection of mild cognitive impairment from spontaneous speech. *Curr Alzheimer Res* 15, 130–138 (2018).

6. National Institute on Aging. Assessing Cognitive Impairment in Older Patients. https://www.nia.nih.gov/health/assessing-cognitive-impairment-older-patients.

7. Meilán, J. J. G., Martínez-Sánchez, F., Martínez-Nicolás, I., Llorente, T. E. & Carro, J. Changes in the rhythm of speech difference between people with nondegenerative mild cognitive impairment and with preclinical dementia. *Behavioural neurology* 2020, (2020).

8. Asgari, M., Kaye, J. & Dodge, H. Predicting mild cognitive impairment from spontaneous spoken utterances. *Alzheimer's & Dementia: Translational Research & Clinical Interventions* 3, 219–228 (2017).

9. Sung, J. E., Choi, S., Eom, B., Yoo, J. K. & Jeong, J. H. Syntactic complexity as a linguistic marker to differentiate mild cognitive impairment from normal aging. *Journal of Speech, Language, and Hearing Research* 63, 1416–1429 (2020).

10. Eyben, F., Scherer, K., … B. S.-I. transactions on & 2015, undefined. The Geneva minimalistic acoustic parameter set (GeMAPS) for voice research and affective computing. *ieeexplore.ieee.orgF Eyben, KR Scherer, BW Schuller, J Sundberg, E André, C Busso, LY Devillers, J EppsIEEE transactions on affective computing, 2015•ieeexplore.ieee.org* https://ieeexplore.ieee.org/abstract/document/7160715/.

11. models/research/audioset/yamnet at master · tensorflow/models. https://github.com/tensorflow/models/tree/master/research/audioset/yamnet.

12. Hershey, S. *et al.* CNN Architectures for Large-Scale Audio Classification. *ICASSP, IEEE International Conference on Acoustics, Speech and Signal Processing - Proceedings* 131–135 (2016) doi:10.1109/ICASSP.2017.7952132.

13. Vaswani, A. *et al.* Attention Is All You Need. *Adv Neural Inf Process Syst* 2017-December, 5999–6009 (2017).

14. Cai, W. *et al.* A Survey on Mixture of Experts. 14, (2024).

15. Majoka, M. A. & Schimming, C. Effect of Social Determinants of Health on Cognition and Risk of Alzheimer Disease and Related Dementias. *Clin Ther* 43, 922–929 (2021).

16. Faisal, F. U. R. & Kwon, G. R. Automated Detection of Alzheimer-s Disease and Mild Cognitive Impairment Using Whole Brain MRI. *IEEE Access* 10, 65055–65066 (2022).

17. Zhao, W. X. *et al.* A Survey of Large Language Models. 한국컴퓨터종합학술대회 논문집 https://arxiv.org/abs/2303.18223v15 (2023).

18. Zolnour, A. *et al.* A risk identification model for detection of patients at risk of antidepressant discontinuation. *Front Artif Intell* 6, 1229609 (2023).

19. Templ, M., Kowarik, A. & Filzmoser, P. Iterative stepwise regression imputation using standard and robust methods. *Comput Stat Data Anal* 55, 2793–2806 (2011).

20. Yang, Q., Li, X., Ding, X., Xu, F. & Ling, Z. Deep learning-based speech analysis for Alzheimer's disease detection: a literature review. *Alzheimers Res Ther* 14, 1–16 (2022).

21. Streijl, R. C., Winkler, S. & Hands, D. S. Mean opinion score (MOS) revisited: methods and applications, limitations and alternatives. *Multimed Syst* 22, 213–227 (2016).

22. Jebb, A. T., Ng, V. & Tay, L. A Review of Key Likert Scale Development Advances: 1995–2019. *Front Psychol* 12, 637547 (2021).

23. Eng., S. A.-Int. J. Comput. Sci. & 2017, undefined. Audio signal noise reduction using low pass filter. *ijcse.netS AroraInt. J. Comput. Sci. Eng., 2017•ijcse.net* http://www.ijcse.net/docs/IJCSE21-10-03-008.pdf.

24. Radford, A. *et al.* Robust speech recognition via large-scale weak supervision. *proceedings.mlr.pressA Radford, JW Kim, T Xu, G Brockman, C McLeavey, I SutskeverInternational conference on machine learning, 2023•proceedings.mlr.press* https://proceedings.mlr.press/v202/radford23a.html.

25. Ali, A. & Renals, S. Word Error Rate Estimation for Speech Recognition: e-WER. *ACL 2018 - 56th Annual Meeting of the Association for Computational Linguistics, Proceedings of the Conference (Long Papers)* 2, 20–24 (2018).

26. Grattafiori, A. *et al.* The Llama 3 Herd of Models. https://arxiv.org/abs/2407.21783v3 (2024).

27. Chen, B., Zhang, Z., Langrené, N. & Zhu, S. Unleashing the potential of prompt engineering in Large Language Models: a comprehensive review. https://arxiv.org/abs/2310.14735v5 (2023).

28. Brown, T. B. *et al.* Language Models are Few-Shot Learners. *Adv Neural Inf Process Syst* 2020-December, (2020).

29. Zhang, Z. *et al.* A Scoping Review of Large Language Model Applications in Healthcare. *Stud Health Technol Inform* 329, 1966–1967 (2025).

30. Detecting Early-Stage Dementia Through Speech - Penn Memory Center. https://pennmemorycenter.org/detecting-early-stage-dementia-through-speech/ (2024).

31. AI screening tool for dementia being developed by Columbia researcher - McKnight's Senior Living. https://www.mcknightsseniorliving.com/news/ai-screening-tool-for-dementia-being-developed-by-columbia-researcher/.

32. Now That's Teamwork: VNS Health Research Center Oct 2024 on Vimeo. https://vimeo.com/1016914299/637b844af7.

33. New Columbia Nursing study on AI tool for detectin | Newswise. https://www.newswise.com/articles/new-columbia-nursing-study-on-ai-tool-for-detecting-alzheimer-s-and-related-dementias-shows-promising-results.

34. Zolnoori, M., Zolnour, A. & Topaz, M. ADscreen: A speech processing-based screening system for automatic identification of patients with Alzheimer's disease and related dementia. *Artif Intell Med* 143, 102624 (2023).

35. J, S., M, Z. & D, S. Do nurses document all discussions of patient problems and nursing interventions in the electronic health record? A pilot study in home healthcare. *JAMIA Open* 5, ooac034.

36. Scroggins, J. K., Topaz, M., Song, J. & Zolnoori, M. Does synthetic data augmentation improve the performances of machine learning classifiers for identifying health problems in patient-nurse verbal communications in home healthcare settings? *J Nurs Scholarsh* https://doi.org/10.1111/JNU.13004 (2024) doi:10.1111/JNU.13004.

37. Topaz, M. *et al.* Speech recognition can help evaluate shared decision making and predict medication adherence in primary care setting. *PLoS One* 17, e0271884 (2022).

38. Zolnoori, M. *et al.* Is the patient speaking or the nurse? Automatic speaker type identification in patient-nurse audio recordings. *J Am Med Inform Assoc* 30, 1673–1683 (2023).

39. Song, J. *et al.* Do nurses document all discussions of patient problems and nursing interventions in the electronic health record? A pilot study in home healthcare. *JAMIA Open* 5, ooac034 (2022).

40. Zolnoori, M. *et al.* Audio recording patient-nurse verbal communications in home health care settings: pilot feasibility and usability study. *JMIR Hum Factors* 9, e35325 (2022).

41. Zolnoori, M. *et al.* Utilizing patient-nurse verbal communication in building risk identification models: the missing critical data stream in home healthcare. *Journal of the American Medical Informatics Association* ocad195 (2023).

42. Zolnoori, M. *et al.* Beyond electronic health record data: leveraging natural language processing and machine learning to uncover cognitive insights from patient-nurse verbal communications. *Journal of the American Medical Informatics Association* ocae300 (2024).

43. Zolnoori, M. *et al.* Decoding disparities: evaluating automatic speech recognition system performance in transcribing Black and White patient verbal communication with nurses in home healthcare. *JAMIA Open* 7, ooae130 (2024).

44. Azadmaleki, H. *et al.* SpeechCARE: Harnessing Multimodal Innovation to Transform Cognitive Impairment Detection – Insights from the National Institute on Aging Alzheimer's Speech Challenge. *Stud Health Technol Inform* 329, 1856–1857 (2025).

45. Zhang, Z. *et al.* Enhancing AI for citation screening in literature reviews: Improving accuracy with ensemble models. *Int J Med Inform* 203, 106035 (2025).

46. Eyben, F., Scherer, K., … B. S.-I. transactions on & 2015, undefined. The Geneva minimalistic acoustic parameter set (GeMAPS) for voice research and affective computing. *ieeexplore.ieee.orgF Eyben, KR Scherer, BW Schuller, J Sundberg, E André, C Busso, LY Devillers, J EppsIEEE transactions on affective computing, 2015•ieeexplore.ieee.org* https://ieeexplore.ieee.org/abstract/document/7160715/.

47. Hsu, W. N. *et al.* HuBERT: Self-Supervised Speech Representation Learning by Masked Prediction of Hidden Units. *IEEE/ACM Trans Audio Speech Lang Process* 29, 3451–3460 (2021).

48. Baevski, A., Zhou, H., Mohamed, A. & Auli, M. wav2vec 2.0: A Framework for Self-Supervised Learning of Speech Representations. *Adv Neural Inf Process Syst* 2020-December, (2020).

49. Babu, A. *et al.* XLS-R: Self-supervised Cross-lingual Speech Representation Learning at Scale. *Proceedings of the Annual Conference of the International Speech Communication Association, INTERSPEECH* 2022-September, 2278–2282 (2021).

50. Boito, M. Z., Iyer, V., Lagos, N., Besacier, L. & Calapodescu, I. mHuBERT-147: A Compact Multilingual HuBERT Model. 3939–3943 (2024) doi:10.21437/interspeech.2024-938.

51. Pires, T., Schlinger, E. & Garrette, D. How Multilingual is Multilingual BERT? *ACL 2019 - 57th Annual Meeting of the Association for Computational Linguistics, Proceedings of the Conference* 4996–5001 (2019) doi:10.18653/V1/P19-1493.

52. Conneau, A. *et al.* Unsupervised Cross-lingual Representation Learning at Scale. *Proceedings of the Annual Meeting of the Association for Computational Linguistics* 8440–8451 (2019) doi:10.18653/v1/2020.acl-main.747.

53. Zhang, X. *et al.* mGTE: Generalized Long-Context Text Representation and Reranking Models for Multilingual Text Retrieval. https://arxiv.org/abs/2407.19669v2 (2024).

54. Chen, J. *et al.* BGE M3-Embedding: Multi-Lingual, Multi-Functionality, Multi-Granularity Text Embeddings Through Self-Knowledge Distillation. https://arxiv.org/abs/2402.03216v4 (2024).

55. Devlin, J., Chang, M. W., Lee, K. & Toutanova, K. BERT: Pre-training of Deep Bidirectional Transformers for Language Understanding. *NAACL HLT 2019 - 2019 Conference of the North American Chapter of the Association for Computational Linguistics: Human Language Technologies - Proceedings of the Conference* 1, 4171–4186 (2018).

56. Gong, Y., Chung, Y. A. & Glass, J. AST: Audio Spectrogram Transformer. *Proceedings of the Annual Conference of the International Speech Communication Association, INTERSPEECH* 1, 56–60 (2021).

57. Agbavor, F. & Liang, H. Predicting dementia from spontaneous speech using large language models. *PLOS digital health* 1, e0000168 (2022).

58. Chen, J.-M. Performance Assessment of ChatGPT vs Bard in Detecting Alzheimer's Dementia. *arXiv preprint arXiv:2402.01751* (2024).

59. Zolnour, A. *et al.* LLMCARE: early detection of cognitive impairment via transformer models enhanced by LLM-generated synthetic data. *Front Artif Intell* 8, 1669896 (2025).

60. mistralai/Ministral-8B-Instruct-2410 · Hugging Face. https://huggingface.co/mistralai/Ministral-8B-Instruct-2410.

61. Dettmers, T., Pagnoni, A., Holtzman, A. & Zettlemoyer, L. QLoRA: Efficient Finetuning of Quantized LLMs. *Adv Neural Inf Process Syst* 36, (2023).

62. Shao, H., Pan, Y., Wang, Y. & Zhang, Y. Modality fusion using auxiliary tasks for dementia detection. *Comput Speech Lang* 95, 101814 (2026).

63. Boulahia, S. Y., Amamra, A., Madi, M. R. & Daikh, S. Early, intermediate and late fusion strategies for robust deep learning-based multimodal action recognition. *Machine Vision and Applications 2021 32:6* 32, 1–18 (2021).

64. Gadzicki, K., Khamsehashari, R. & Zetzsche, C. Early vs late fusion in multimodal convolutional neural networks. *Proceedings of 2020 23rd International Conference on Information Fusion, FUSION 2020* https://doi.org/10.23919/FUSION45008.2020.9190246 (2020) doi:10.23919/FUSION45008.2020.9190246.

65. Wu, Z., Gong, Z., Koo, J. & Hirschberg, J. Multimodal Multi-loss Fusion Network for Sentiment Analysis. *Proceedings of the 2024 Conference of the North American Chapter of the Association for Computational Linguistics: Human Language Technologies, NAACL 2024* 1, 3588–3602 (2024).

66. Poor, F. F., Dodge, H. H. & Mahoor, M. H. A multimodal cross-transformer-based model to predict mild cognitive impairment using speech, language and vision. *Comput Biol Med* 182, 109199 (2024).

67. Hardt, M., Price, E., information, N. S.-A. in neural & 2016, undefined. Equality of opportunity in supervised learning. *proceedings.neurips.ccM Hardt, E Price, N SrebroAdvances in neural information processing systems, 2016•proceedings.neurips.cc* https://proceedings.neurips.cc/paper/2016/hash/9d2682367c3935defcb1f9e247a97c0d-Abstract.html.

68. Xu, Z. *et al.* Voice for All: Evaluating the Accuracy and Equity of Automatic Speech Recognition Systems in Transcribing Patient Communications in Home Healthcare. *Stud Health Technol Inform* 329, 1904–1906 (2025).

69. Park, D. S. *et al.* SpecAugment: A Simple Data Augmentation Method for Automatic Speech Recognition. *Proceedings of the Annual Conference of the International Speech Communication Association, INTERSPEECH* 2019-September, 2613–2617 (2019).

70. Walczyna, T. & Piotrowski, Z. Overview of Voice Conversion Methods Based on Deep Learning. *Applied Sciences 2023, Vol. 13, Page 3100* 13, 3100 (2023).

71. Zhang, B. H., Lemoine, B. & Mitchell, M. Mitigating Unwanted Biases with Adversarial Learning. *AIES 2018 - Proceedings of the 2018 AAAI/ACM Conference on AI, Ethics, and Society* 335–340 (2018) doi:10.1145/3278721.3278779.

72. Lin, T. Y., Goyal, P., Girshick, R., He, K. & Dollar, P. Focal Loss for Dense Object Detection. *IEEE Trans Pattern Anal Mach Intell* 42, 318–327 (2017).

73. Rashidi, S., Azadmaleki, H., Zolnour, A., Momeni Nezhad, M. J. & Zolnoori, M. SpeechCura: A Novel Speech Augmentation Framework to Tackle Data Scarcity in Healthcare. *Stud Health Technol Inform* 329, 1858–1859 (2025).